# Symmetry Breaking with Polynomial Delay

Tim Januschowski

Cork Constraint Computation Centre

Computer Science Department, UCC, Ireland

Barbara M. Smith

School of Computing, University of Leeds, U.K.

M. R. C. van Dongen

Computer Science Department, UCC, Ireland

May 30, 2018

## Abstract

A *conservative* class of constraint satisfaction problems (CSPs) is a class for which membership is preserved under arbitrary domain reductions. Many well-known tractable classes of CSPs are conservative. It is well known that lexleader constraints may significantly reduce the number of solutions by excluding symmetric solutions of CSPs. We show that adding certain lexleader constraints to any instance of any conservative class of CSPs still allows us to find all solutions with a time which is polynomial between successive solutions. The time is polynomial in the total size of the instance and the additional lexleader constraints. It is well known that for complete symmetry breaking one may need an exponential number of lexleader constraints. However, in practice, the number of additional lexleader constraints is typically polynomial number in the size of the instance. For polynomially many lexleader constraints, we may in general not have complete symmetry breaking, but polynomially many lexleader constraints may provide practically useful symmetry breaking – and they sometimes exclude super-exponentially many solutions. We prove that for any instance from a conservative class, the time between finding successive solutions of the instance with polynomially many additional lexleader constraints is polynomial even in the size of the instance *without* lexleader constraints.

# 1 Introduction

In many applications of constraint satisfaction problems (CSPs), it often does not suffice to find only one solution, but the task is to find all solutions. Examples range from problems of theoretical interest (see e.g. Distler and Kelsey, 2009; Behle and Eisenbrand, 2007; Bussieck and Lübbecke, 1998) to industrial applications (see e.g. Shlyakhter, 2007; Ip and Dill, 1996, and references therein). The task of finding all solutions of a CSP is #P-complete, which is much harder than the already hard task of finding one solution, which is $\mathcal{NP}$-complete Garey and Johnson (1979). Even for those classes where we can prove satisfiability in polynomial time, it may be $\mathcal{NP}$-complete to decide whether a second solution exists. For example, consider the following class of CSPs. Given any CSP, we create a new value $v$ and add it to the domain of each variable. Next we modify the constraints such that it allows one more assignment which assigns $v$ to each variable. For this class of CSPs, the problem of finding one solution is polynomial. However, deciding whether the CSP has a second solution is $\mathcal{NP}$-complete. Identifying classes of CSPs where we can find all solutions in an "acceptable time" has therefore received considerable attention (see e.g. Schnoor and Schnoor, 2007; Cohen, 2004; Greco and Scarcello, 2010). Acceptable time for finding all solutions formalises as follows. First, deciding the satisfiability of any instance of the class is polynomial. Finally, the class has an algorithm which, for any instance in the class, requires (1) polynomial time between start and the first solution, (2) polynomial time between successive solutions, and (3) polynomial time between the last solution and termination. The class is then said to have *polynomial delay*.

Symmetries are a frequent feature of many practically important CSPs. Symmetries of a CSP are permutations of the variable-value assignments that preserve the constraints. The presence of symmetries indicates a redundancy in the CSP: symmetries partition the set of solutions into disjoint classes of symmetrically equivalent solutions. For symmetric CSPs, it is often enough to only find all equivalence classes of solutions as opposed to finding all solutions (see e.g. Shlyakhter, 2007; Ip and Dill, 1996, and references therein). Finding equivalence classes of solutions is particularly interesting if the symmetries of the CSP are model-introduced (see e.g. Smith, 2006; Frisch et al., 2005). Finding all equivalence classes of symmetric solutions can be achieved by adding symmetry breaking constraints to the CSP, which exclude symmetric solutions while keeping at least one representative solution per equivalence class. In this paper, we consider adding symmetry breaking constraints to CSPs with polynomial delay and we study the complexity of the resulting class.

**Main Contribution.** We show that excluding symmetric solutions by adding certain lexleader constraints (Crawford et al., 1996) *preserves polynomial delay* for many well-known tractable classes which are *conservative*. Here a class is conservative if it is closed under *arbitrary domain reductions*, a notion which we formally define further on. Bulatov (2003) defines conservative constraint languages, but our definition is more general. We prove that symmetry breaking is possible with a delay which is polynomial in the size of the instance and the size of the lexleader constraints. This includes complete symmetry breaking (with possibly super-exponentially many lexleader constraints). However, if symmetry is broken by adding a polynomial number of lexleader constraints, then the delay is polynomial in the size of the instance *only*. Of course, such symmetry breaking may not be complete.

This is a non-trivial result as we argue in the following. A naïve approach would simply generate all solutions regardless of the lexleader constraints and, after generation, reject solutions when they are symmetrically equivalent to an already found one. This approach would not preserve polynomial delay because the number of solutions rejected between successive symmetrically-distinct solutions can be super-exponential. A simple example shows this. Consider a CSP with $n$ variables $x_1, x_2, x_3, \ldots, x_n$. Each variable has domain $\{1, 2, 3, \ldots, n\}$ and we have one alldifferent constraint (decomposed into binary $\neq$ constraints). A straight-forward extension of a result by Salamon and Jeavons (2008) yields that this CSP has polynomial delay. The variable symmetries can be broken completely by adding a linear number of constraints $x_i \leq x_{i+1}$ for $i \in \{1, 2, \ldots, n-1\}$(Puget, 2005; Grayland et al., 2009). Without the symmetry breaking constraints, there are $n!$ many solutions and with the symmetry breaking constraints there is 1 solution. The afore-mentioned generate-and-test approach would not have polynomial delay. Here, we prove that adding lexleader constraints nevertheless has the potential of finding all symmetrically distinct solutions of a CSP with polynomial delay even though we may sometimes exclude a super-exponential number of solutions with a polynomial number of lexleader constraints. In particular, we show that the delay of the CSP with lexleader constraints is polynomial even in the size of the instance without lexleader constraints, if we limit the number of lexleader constraints to a polynomial number. So, we can find all equivalence classes of solutions much faster with the help of lexleader constraints.

**Structure of the Paper.** The organisation of this paper is as follows. In the first part, we consider lexleader constraint in isolation from other constraints and in the second part, we consider them in combination with problem-specific constraints.

When considering lexleader constraints in isolation from problem-specific constraints, we start by introducing the class LEX of lexleader constraints on which we focus our attention in this paper in Section 4.1. In the same section, we also consider CSPs whose constraints are in LEX. We shall refer to this class of CSPs as LCSP. We show that LCSP has polynomial delay. To the best of our knowledge, this result does not follow from known results on classes with polynomial delay.

When considering lexleader constraints in combination with problem-specific constraints, we present an algorithm in Section 5 that finds all solutions of a CSP with lexleader constraints in LEX. We prove that the algorithm requires polynomially many calls to an oracle between finding subsequent solutions. This is our main result. Since the oracle runs in polynomial time for many well-known classes with polynomial delay, the preservation of polynomial delay under the addition of lexleader constraints follows directly.

We start by introducing notation and discussing related work.

## 2  Notation and Definitions

A *constraint satisfaction problem* (CSP) is a triple $C = (\mathcal{X}, \mathcal{D}, \mathcal{C})$, where $\mathcal{X}$ is the set of variables of $C$, every variable $x$ has a *domain* $\mathcal{D}(x) \in \mathcal{D}$, and $\mathcal{C}$ is the set of constraints of $C$. Every constraint has an *arity*. The $k$-ary constraint $c$ is a pair $\langle s, r \rangle$, where $s$ is a list of $k$ variables $x_1, \ldots, x_k$ which is called the *scope* and $r$ is called *the relation* of $c$. The relation is either given extensionally as a list of tuples that is allowed by the constraint ($r \subseteq \mathcal{D}(x_1) \times \cdots \times \mathcal{D}(x_k)$) or intensionally as an expression. The arity of $C$ is the maximum arity over all constraints in $C$. A CSP is called *binary* if all constraints are of arity 2 or lower. A *literal* is a (variable,value)-assignment. A *partial assignment* is a set of literals such that no variable appears twice. If a partial assignment is allowed by the constraints of $C$, then we call it a *consistent assignment*. A *solution* is a consistent assignment on all variables. If $C$ has a solution, then we call $C$ *satisfiable* and otherwise *unsatisfiable*. For a CSP with $n$ variables, we denote an *order* $\prec$ on the variables either as a list $[x_1, x_2, \ldots, x_n]$ or by stating $x_1 \prec x_2 \prec \cdots \prec x_n$. We call a partial assignment to the first variables with respect to $\prec$ a *consecutive partial assignment with respect to $\prec$*. We assume from here on, that the domain of any variable $x$ contains all integers between two values min and max, denoted by $[\min, \max]$. We refer to the smallest value that we can consistently assign to a variable as the *lower bound* of the variable.

We call a constraint $c = \langle s, r \rangle$ *generalised arc consistent* (GAC) if the following holds for all variables in $s = [x_1, x_2, \ldots, x_k]$. For any $x_i \in s$ and $d_i \in \mathcal{D}(x_i)$ with $i \in \{1, \ldots, k\}$, there is a *support*, that is an assignment $\{(x_1, d_1), (x_2, d_2), \ldots, (x_k, d_k)\} \in r$ such that $d_j \in \mathcal{D}(x_j)$ for $\{1, \ldots k\} \ni j \neq i$. If the constraint is binary and GAC, we simply say it is *arc consistent*. We can make a constraint GAC by removing those values from the domains of the variables in the scope of the constraint which do not have support.

Let us define the *size* of a CSP. Given an instance $(\mathcal{X}, \mathcal{D}, \mathcal{C})$, we define the size of the extensional constraints first. We represent each constraint scope as a list of variables $s \subseteq \mathcal{X}$, and each constraint relation as a list of tuples over $\mathcal{D}(x)$, $x \in s$. The size of the extensional constraints is then

$$\log(|\mathcal{X}|) + \sum_{\langle s,r \rangle \in \mathcal{C}} |s||r| \log(M),$$

where $M = \max_{x \in \mathcal{X}} |\mathcal{D}(x)|$. In this paper, we only consider one type of intensional constraint, which we introduce in Section 3. The size of an intensional constraint of this type is $(n \log(|\mathcal{X}|))$. The size of a CSP with constraints in both extensional and intensional form is the sum of the size of the extensional constraints with the size of the intensional constraints.

A class of CSPs is said to be *tractable* if we can decide the satisfiability of any instance in this class in a time that is polynomial in the size of the instance. A tractable class is said to have *tractable search* if for any member of the class that has a solution, we can find a solution in polynomial time. A tractable class of CSPs has *polynomial delay* if for any instance in the class, the following times are polynomial in the size of the instance: the time between the start and the first solution, the time between consecutive solutions and the time between the last solution, and the termination of the algorithm.

We call a class of CSPs *conservative* if class membership is preserved under domain reductions (Cooper et al., 2010). A *domain reduction* of a domain $\mathcal{D}$ is either the replacement of $\mathcal{D}$ by $\mathcal{D}' \subseteq \mathcal{D}$ together with adjustments of the constraints (e.g. removing tuples from constraint relations); or a domain reduction is the addition of a unary constraint that reduces the size of the domain of a variable.

With any CSP $C$ we associate a hypergraph called the *microstructure complement* (MSC). The set of nodes of the MSC are the literals of $C$. A set of nodes $a = \{(x_1, d_1), \ldots, (x_k, d_k)\}$ is a hyperedge in the MSC either if $k = 2$ and $x_1 = x_2$ or if $a$ is forbidden by a constraint of $C$. An *automorphism* of a hypergraph is a bijection on the set of nodes that preserves hyperedges. A *constraint symmetry* (Cohen et al., 2006) of $C$ is an automorphism of the MSC of $C$. A *variable constraint symmetry* is a constraint symmetry $\phi$ such that $\phi(x, d) = (\psi(x), d)$ for a suitable chosen bijective map $\psi$ on the variables of $C$. Constraint symmetries form a group and partition the set of solutions of a CSP into disjoint *orbits*, which are equivalence classes of symmetric solutions.

## 3 Related Work

Apart from constraint symmetries, other forms of symmetries exist, most notably *solution symmetries*. Solution symmetries are a supergroup of constraint symmetries and our results hold for solution symmetries as well. However, solution symmetries are typically very hard to find and, in practice, Constraint Programmers consider constraint symmetries (Cohen et al., 2006). Two frequent constraint symmetries are variable constraint symmetries and value constraint symmetries. A value constraint symmetry $\phi: \mathcal{X} \times \mathcal{D} \to \mathcal{X} \times \mathcal{D}$ is a constraint symmetry such that $\phi(x, d) = (x, \psi(d))$ for a suitably chosen bijective map $\psi: \mathcal{D} \to \mathcal{D}$ on the values. We can ensure in polynomial time that in a search tree as constructed by a backtrack algorithm no two nodes are symmetric via a value symmetry (Roney-Dougal et al., 2004). Hence, results on the preservation of polynomial delay naturally hold for problems with value constraint symmetry breaking. However, for variable constraint symmetries, which we consider here, results on the preservation of polynomial delay are new.

In this paper, we consider adding symmetry breaking constraints to a CSP *before* solving the CSP. This is usually referred to as *static symmetry breaking*. Other symmetry breaking methods add symmetry breaking constraints during search dynamically. See (Gent et al., 2006) for an overview of general approaches to symmetry breaking in constraint programming. For static symmetry breaking, Puget (1993) presented an abstract framework and provided a basic symmetry breaking constraint. Crawford et al. (1996) then introduced general symmetry breaking constraints in the context of Boolean satisfiability. We consider static symmetry breaking for variable constraint symmetries. A variable constraint symmetry $\sigma$ of a CSP with $n$ variables $x_1, \ldots, x_n$ can be thought of as a permutation of the index set $\{1, 2, \ldots, n\}$. Let us choose an order on the variables, say $[x_1, x_2, ..., x_n]$.

A lexleader constraint (Crawford et al., 1996) enforces that any consistent assignment on the variables $[x_1, x_2, ..., x_n]$ does not lexicographically exceed its symmetric image under $\sigma$. In formula, the lexleader constraint is

$$[x_1, x_2, \ldots, x_n] \leq_{lex} [x_{\sigma(1)}, x_{\sigma(2)}, \ldots, x_{\sigma(n)}].$$

We represent lexleader constraints intensionally because their extensional representation may quickly lead to exponential space requirement. We refer to the variables on the left hand side of $\leq_{lex}$ as LHS and to the variables on the right hand side as RHS. For lexicographic comparison, we also need an order on the domains of the variables. In this paper, we shall use the natural order on the integers. We refer to the variable order in the lexleader constraint and the order on the domains as the *orders of the lexleader constraints*. Lexleader constraints preserve *at least one* solution from every orbit of solutions (Crawford et al., 1996). We call this property *partial symmetry breaking*. If we introduce a lexleader constraint for every symmetry, then we preserve *exactly one* solution per orbit (Crawford et al., 1996). We call this property *complete symmetry breaking*. Unfortunately, the number of symmetries can be super-exponential in the number of variables. This is why practitioners typically do not add all possible lexleader constraints to a CSP. Next, we discuss two approaches to simplify lexleader constraints.

There are two main approaches to reducing the number and arity of lexleader constraints which we present next. The first approach is partial symmetry breaking, that is, we only construct lexleader constraints for a subset of the symmetries (see e.g. Gent et al., 2006). The second approach consists of reduction rules that simplify a set of lexleader constraints while preserving their logical equivalence.

An example of partial symmetry breaking is any set of lexleader constraints based on *transpositions*. A transposition is a permutation that swaps two variables while having no effect on the other variables. There are at most $\binom{n}{2}$ transpositions of $n$ variables, although the symmetry *group* can be much larger: for instance if every transposition is a symmetry, then every permutation of the variables is a symmetry, and the size of the group is $n!$. A polynomially-sized set of lexleader constraints does not always provide complete symmetry breaking and often only removes a polynomial number of solutions. However, in some cases, a polynomially-sized set of lexleader constraints does remove an exponential number of solutions: whenever the symmetry group contains all transpositions, adding lexleader constraints based on transpositions break the symmetries completely (see e.g. Grayland et al., 2009; Puget, 2005). In many cases, partial symmetry breaking may be a good alternative from a user's point of view (see e.g. Shlyakhter, 2007; Flener et al., 2002; Grayland et al., 2009).

We briefly discuss an example of a reduction rule which reduces the arity of lexleader constraints (see e.g. Grayland et al., 2009). In Section 4.1, we shall consider a family of lexleader constraints which subsumes lexleader constraints that are reduced with this rule. Suppose that we are not deriving lexleader constraints for all symmetries, but only for symmetries of a CSP with $n$ variables $x_1, \ldots, x_n$ that can be written as a composition of disjoint transpositions. The advantage of a lexleader constraint based on a composition of disjoint transpositions $\sigma$ is that we can easily simplify the resulting lexleader constraints to give lexleader constraints without repeated variables, as follows.

- If $\sigma(i) = i$, then $x_i$ can be dropped from both sides of the lexleader constraint;
- For indices $a$ and $b$ with $a < b$, if $\sigma$ swaps the indices $a$ and $b$, then the constraint will first have $x_a$ on the LHS, paired with $x_b$ on the RHS, and then later $x_b$ on the LHS and $x_a$ on the RHS. We can drop the second pair. If we have $x_i < x_{\sigma(i)}$ for some $i$ such that $1 \leq i < b$, then $x_a < x_b$ and the second pair has no effect because the assignment on LHS is already lexicographically less than the assignment on RHS. If $x_i = x_{\sigma(i)}$ for $1 \leq i < b$, then in particular $x_a = x_b$ and again, the second pair has no effect.

This is an example of a reduction rule for lexleader constraints.

We consider adding symmetry breaking constraints to classes of CSPs with polynomial delay. Cohen (2004) shows that certain classes defined by restrictions on the constraint relations have polynomial delay via an algorithm which we modify for our purposes further on. Schnoor and Schnoor (2007) provide extensions of Cohen's results and present new enumeration schemes for finding all solutions. Greco and Scarcello (2010) prove the polynomial delay of a wide range of classes defined by restrictions on constraint scopes. However, none of their work considers symmetry breaking. We are not aware of work on problems with polynomial delay with added symmetry breaking constraints.

To the best of our knowledge, no theoretical results on the effect of symmetry breaking on tractability or polynomial delay are known. Empirical results do not provide a clear indication on the effect of symmetry breaking on tractability. For example, there is abundant evidence of the practical usefulness of symmetry breaking (see e.g. Gent et al., 2006). However, evidence also shows that symmetry breaking may be harmful sometimes (see e.g. Gent et al., 2002; Prestwich, 2008).

## 4 CSPs Consisting of Lexleader Constraints

This section studies lexleader constraints in isolation from problem-specific constraints. We introduce LCSP as a class of CSPs that has only lexleader constraints of a certain form and we show that LCSP has polynomial delay. Our results on combining lexleader constraints with other constraints build on this result.

### 4.1 Finding All Solutions

In this section, we prove that for lexleader constraints of a certain form, we can find all solutions of a CSP whose constraints consist of these lexleader constraints with failure-free search. We use this result to show that the class of CSPs with these lexleader constraints has polynomial delay. First, we define the class of lexleader constraints that we need for the exposition of our results.

Consider a CSP with $n$ variables ordered as $[x_1, x_2, ..., x_n]$. Let us define the subset of lexleader constraints that we want to study in the following.

**Definition 1** (LEX). *Let $\mathcal{X}$ be a set of variables ordered as $[x_1, x_2, ..., x_n]$. A family of lexleader constraints $L$ on $\mathcal{X}$ is in LEX if each lexleader constraint in $L$ has the following characteristics.*

- *The constraint is of the form*

$$[x_{\ell_1}, x_{\ell_2}, x_{\ell_3}, \ldots, x_{\ell_k}] \leq_{lex} [x_{r_1}, x_{r_2}, x_{r_3}, \ldots, x_{r_k}]. \qquad (1)$$

  *Often, we consider variables in LHS and RHS with the same sub-index $i$ and refer to them as a pair $(x_{\ell_i}, x_{r_i})$ with respect to Constraint (1). We omit references to the lexleader constraint whenever it is clear what we mean.*
- *For each $j$ with $1 \leq j < k$, we have $\ell_j < \ell_{j+1}$.*
- *For each $j$ with $1 \leq j \leq k$, we have $\ell_j \leq r_j$.*

Lexleader constraints for compositions of disjoint transpositions can be reduced (with the reduction rule discussed in Section 3) such that they are contained in LEX. Let us define a class of CSPs that we want to study in the following.

**Definition 2** (LCSP). *We denote the class of CSPs consisting of constraints in LEX by LCSP.*

The following theorem explains the effect of maintaining GAC during search for solutions of CSPs in LCSP with a chronological backtrack algorithm that maintains GAC.

**Theorem 3.** *For any CSP in LCSP, where we search for a solution with a chronological backtrack search with variable and value orders in the search being the same as in the lexleader constraints, maintaining GAC may only result in an increase in the lower bound on the unassigned variables.*

*Proof.* Suppose that at some point in the search the variables $x_1, x_2, \ldots, x_i$ have been assigned during search and the remainder have not. We show in the following that maintaining GAC may only remove the smallest values in the domains of the unassigned variables.

We consider an unassigned variable. Let us first consider the case, where an unassigned variable occurs on the RHS of some constraint, paired with a LHS variable that is already assigned. We refer to the RHS variable as $x_{r_j}$ and the LHS variable as $x_{\ell_j}$, where $x_{\ell_j}$ has been assigned the value $a$, for some value $\min \leq a \leq \max$; $\ell_j \leq i$ and $r_j > i$. Let us consider the effect of constraint propagation on the domain of such a variable $x_{r_j}$. First, all the preceding variables on both sides of the constraint may have been assigned such that the LHS and RHS variable in each pair are equal. In that case, we have $x_{r_j} \geq a$. Next, a pair of variables in the same constraint could exist that appears behind $(x_{\ell_j}, x_{r_j})$ in the order and the pair is such that the RHS variable must have a smaller value than the corresponding LHS variable. The pair $(x_{\ell_j}, x_{r_j})$ could then be the only pair of variables that will allow the constraint to be satisfied, by having $x_{\ell_j} < x_{r_j}$, resulting in $x_{r_j} \geq a+1$. (Note that because GAC is maintained on this constraint, it must be possible to do this. We must have $x_{\ell_j} \leq x_{r_j}$, so if $x_{\ell_j}$ is assigned the value max, then the domain of $x_{r_j}$ becomes max and it would not be possible to assign the variables that appear later in the order in the constraint in a way that forces $x_{r_j} > x_{\ell_j}$.) Hence, if a variable occurs on the RHS of a constraint, propagating the constraint results in an increase of the lower bound.

Next, let us consider the case where an unassigned variable $x_{\ell_j}$ appears on the LHS of some constraint. We have $\ell_j > i$. If $x_{\ell_j}$ appears on the RHS of a constraints, its lower bound may have increased as argued before. So in each constraint, when the LHS variables have not yet been assigned, any variable appearing later in the order either in LHS or RHS may have had its lower bound increased as a result of enforcing GAC on other constraints. In the extreme case, enforcing GAC on another constraint may result in $x_{\ell_j} = \max$. Because of the previous domain reductions we must have $x_{\ell_j} \leq x_{r_j}$, so $x_{r_j}$ and every later variable on both sides of the constraint must be instantiated to max.

To summarise, the effect of maintaining GAC on a set of lexleader constraints on the domain of any of the variables $x_{i+1}, x_{i+2}, ..., x_n$ can only be that the lower bound is increased. In particular the value max is not removed. □

**Corollary 4.** *For any CSP in LCSP, we can find all solutions failure-free, if we maintain GAC on the lexleader constraints during search.*

*Proof.* In a chronological backtrack algorithm, we choose the variable and value order in the search as the same as in the lexleader constraints. In light of Theorem 3, we only have to prove that we can extend any consistent assignment to a solution. This follows from the fact that value max is not removed by maintaining GAC: in a consistent partial assignment as found by the chronological backtrack algorithm, we can assign max to all unassigned variables which is consistent with all lexleader constraints. □

Theorem 3 shows that when assigning values to the variables in an order that is compatible with the lexleader constraints, maintaining GAC only removes values from the lower end of any domain. If different orders are used for search and for constructing the lexleader constraints, then failures may occur, as we show in the following example.

**Example 5.** *Suppose all variables in the CSP have domain $[0,1]$ and the lexleader constraints constructed for order $[x_1, x_2, \ldots, x_6]$ are*

$$[x_1, x_5] \leq_{lex} [x_2, x_6] \tag{2}$$

*and*

$$[x_1, x_2, x_3] \leq_{lex} [x_5, x_6, x_4]. \tag{3}$$

*Note that these constraints are in LEX. Suppose we make the assignments: $x_1 = 0$, $x_2 = 0$, $x_3 = 1$ and $x_4 = 0$. Since $x_1 = x_2$, it follows from (2) that $x_5 \leq x_6$. Since $x_3 > x_4$, it follows from (3) that either $x_5 > x_1 = 0$ or $x_6 > x_2 = 0$. Enforcing GAC on each constraint reduces the domain of neither $x_5$ nor $x_6$.*

*Following Theorem 3, we should assign $x_5$ next. Either value in the domain of $x_5$ will lead to a solution. In each case the value 0 will be removed from the domain of $x_6$. One can confirm easily that we can find all solutions failure-free, if we use the orders as suggested in Theorem 3.*

*Consider using an order that is not in accordance with the order in the lexleader constraints and hence not fulfilling the condition of Theorem 3: instead of assigning $x_5$, we try to assign a value to $x_6$. If we try to assign the value* 0, *the assignment fails after* GAC *and we need to backtrack.*

This example shows that assigning the variables in a different order than for the lexleader constraints may lead to failures during search. However, the example confirms that all solutions can be found without failing if the correct variable solutions can be found without failing if the correct variable and value orders are followed. Empirically, it has been known for a long time that lexleader constraints may increase the run-time as opposed to reducing it (Gent et al., 2002). Also well-known is the advice to use the same orders for search and for the construction of lexleader constraints (see e.g. Gent et al., 2006). Theorem 3 gives further theoretical justifications for using the same orders.

The following result is the main result of this section.

**Theorem 6.** *LCSP has polynomial delay.*

*Proof.* Consider a chronological backtrack algorithm with the variable and value order of the lexleader constraints. We have to prove that the time from the start of the chronological backtrack algorithm to the first solution, the time between consecutive solutions and the time between the last solution and termination of the algorithm is polynomial.

The time from start to the first solution is polynomial, because assigning min to every variable leads to (the first) solution and the time from last solution to termination is also polynomial, because assigning max to every variable is allowed (and the last solution).

Hence, we only prove that the time between consecutive solutions is polynomial. Consider a node in the search tree and the set of lexleader constraints. We can enforce GAC on a single lexleader constraints in polynomial time (Kiziltan, 2004) and we can also ensure in polynomial time that each lexleader constraints is GAC.[1] Corollary 4 guarantees that the current subtree contains a solution which we can then find in polynomial time. □

We note that Theorem 6 also holds for the practically important case where there is only a polynomial number (in the variables) of lexleader constraints in LCSP. We further remark that for sets of lexleader constraints in LCSP that have super-exponential size in the number of the variables, a simple generate-and-test approach is enough to prove Theorem 6.

---

[1]This is in contrast to achieving GAC on a conjunction of lexleader constraints for which enforcing GAC is $\mathcal{NP}$-hard (Bessière et al., 2004).

In the next section, we prove stronger results on combining lexleader constraints in LEX with problem-specific constraints.

## 5   Adding Lexleader Constraints to Symmetric CSPs

In the previous section, we considered CSPs with lexleader constraints in isolation from problem-specific constraints. In practice, we consider CSPs with lexleader constraints in combination with other constraints. This is the setting of this section.

Before we start, we state two assumptions. Our first assumption is that we ignore the problem of finding symmetries. In many cases, the symmetries are provided by the user through their insight into a specific problem (see Gent et al., 2006). For automated symmetry detection, finding symmetries is as hard as the graph isomorphism problem (Crawford, 1992) whose complexity status is a long-standing open problem (Garey and Johnson, 1979). It is well-solved in practice (see e.g. McKay, 1981). Our second assumption is that we ignore the complexity of reducing lexleader constraints such that they fit the class LEX. For example, lexleader constraints for symmetries that are disjoint compositions of transpositions can easily be reduced. Note, that it may not always be possible to reduce lexleader constraints such that they fit into LCSP, however, in practice, disjoint transpositions occur frequently (e.g. row-and column symmetries in matrix models Flener et al., 2002).

For this section, we assume the existence of an oracle, that tells us for any given CSP with an order on the variables whether a consecutive partial assignment (with respect to this order) can be extended to a solution. We simply refer to this as *the* oracle. We provide Algorithm 1 that finds all solutions to any CSP where we have added lexleader constraint in LEX. The solutions that Algorithm 1 outputs are the solutions of the CSP modulo the lexleader constraints: these are usually much fewer solutions than the CSPs without lexleader constraints has. In Theorem 7 we prove that Algorithm 1 works with polynomial delay in calls to the oracle. We call this *polynomial oracle-delay.* Since the oracle can decide the satisfiability of any CSP easily, we do not further discuss the case where the input CSP is unsatisfiable.

In the following, we describe Algorithm 1. Algorithm 1 is a chronological backtrack search for all solutions similar to algorithm `PolEnum` (Cohen, 2004). We assume that every variable $x_i$ has a domain $D_i$ that we reduce during search by maintaining GAC. Maintaining GAC is the main difference between our algorithm and `PolEnum`. We assume that domains are backtrackable, i.e., domains restore themselves upon backtracking. Given a CSP $C$, we fix an order in Step 1 which we use both for search and the lexleader constraints. In order to facilitate exposition, we say that this step constructs a second CSP $C_{\text{LEX}}$ which consists of the same variables and domains as $C$. The constraints of $C_{\text{LEX}}$ are in LEX for the specified order and so, $C_{\text{LEX}} \in \text{LCSP}$. We search for all solutions in $C_{\text{LEX}}$ with a chronological backtrack search with a fixed variable and value order as in Corollary 4 maintaining GAC on the lexleader constraints (Step 10). The assignment step of the algorithm is in two stages. First, we tentatively assign the smallest allowed value to the next variable in the search order in $C_{\text{LEX}}$ and propagate. This step will always succeed. Next, we turn to the oracle in Step 11 and ask whether the current partial assignment can be extended to a solution in $C$. The oracle takes all domain reductions through enforcing GAC into account. If the oracle confirms the existence of a solution in $C$, we consider the next variable in the search order; if no solution exists, we consider the next value, removing the current value from the domain of the variable (Step 18). We propagate the effect of the oracle call using $C_{\text{LEX}}$. We backtrack in Step 21 and restore the domains to the state just before the undone assignment happened.

The following result shows that Algorithm 1 finds solutions with polynomial oracle-delay.

**Theorem 7.** *Algorithm 1 finds solutions with polynomial oracle-delay.*

*Proof.* Given a CSP $C$ as input to the algorithm, let the CSP $C_{\text{LEX}}$ consist of the variables and domains of $C$ and lexleader constraints in LEX for the input symmetries. We assign values to variables in such a way that the only effect that enforcing GAC on the lexleader constraints in $C_{\text{LEX}}$ has, is a raise in the lower bound of the unassigned variables. This is analogous to Theorem 3. As opposed to the proof of Corollary 4, however, we cannot simply set all remaining variables to max to have a guaranteed solution for the branch of the search tree (as this could be forbidden by the constraints in $C$). We have to prove that the current assignment on the first $i$ variables can be extended to a solution in $C$ that also satisfies the constraints in $C_{\text{LEX}}$. For the sake of contradiction, we assume that the current assignment cannot be extended to a solution that satisfies all constraints in $C_{\text{LEX}}$.

The oracle guarantees that the current assignment can be extended to a solution $S$ of $C$. If $S$ is forbidden by a constraint in $C_{\text{LEX}}$, then a symmetry $\phi$ exists with $\phi(S) <_{lex} S$. So, $\phi(S)$ has a strictly smaller value $v$ on the first position where $S$ and $\phi(S)$ differ, say at position $j$. If $j > i$, then there

**Input**: CSP $C$ with $n$ variables, each variables has domain $\mathcal{D}$ and a set $\mathcal{S}$ of symmetries of $C$ such that the corresponding lexleader constraints for $\mathcal{S}$ are in LEX ;
**Output**: All symmetrically distinct solutions of $C$ with respect to $\mathcal{S}$

1. Fix search order $[x_1, x_2, \ldots, x_n]$ ;
2. Construct lexleader constraints in LEX with order $[x_1, x_2, \ldots, x_n]$ ;
3. For each variable $x_i$ define a backtrackable domain $\mathcal{D}_i = \mathcal{D}$ ;
4. Define a partial assignment $Assign = \emptyset$ ;
5. $i \leftarrow 1$ ;
6. **while** $i > 0$ **do**
7.     **if** $\mathcal{D}_i \neq \emptyset$ **then**
8.         $d \leftarrow \min(\mathcal{D}_i)$ ;
9.         $Assign \leftarrow Assign \cup \{(x_i, d)\}$ ;
10.         Enforce GAC on lexleader constraints ;
11.         **if** *Assign extends to a solution in $C$* **then**
12.             **if** $i = n$ **then**
13.                 Print *Assign* as the next solution ;
14.             **else**
15.                 $i \leftarrow i + 1$ ;
16.         **else**
17.             $Assign \leftarrow Assign \setminus \{last\ assignment\ to\ x_i\}$ ;
18.         $\mathcal{D}_i \leftarrow \mathcal{D}_i \setminus \{d\}$ ;
19.     **else**
        // backtrack
20.         $\mathcal{D}_i \leftarrow \mathcal{D}$ ;
21.         $Assign \leftarrow Assign \setminus \{last\ assignment\ to\ x_i\}$ ;
22.         $i \leftarrow i - 1$ ;

**Algorithm 1:** Chronological Backtrack for Finding All Solutions.

is a solution in the current branch, which we assumed to be impossible. So, $j \leq i$. However, value $v$ for variable $j$ was removed by enforcing GAC on the lexleader constraint for symmetry $\phi$, when finding solution $S$, because we always choose the smallest available value. Hence, the oracle should have indicated that the current assignment cannot be extended to a solution on the reduced domains.

This proves that for the current branch of the search tree, we can surely find a solution. Hence, we can use the same argument as in the proof of Theorem 6 to show that the time between finding consecutive solutions and the time to find the first solution is polynomial. To terminate the proof, we only have to show that the time between the last solution and the termination of the algorithm requires a polynomial number of calls to the oracle. We have already shown that whenever the oracle confirms that a solution still exists for the current branch, a solution in this branch exists that is allowed by the lexleader constraints as well. So, in the worst case, we have to backtrack to the last assigned variable and call the oracle for all remaining values for the variable until we have reached the first variable. Hence, a polynomial number of calls to the oracle suffices to terminate the algorithm. □

In the remainder of this section, we use Algorithm 1 and Theorem 7 to show that for some tractable classes of CSPs, we can find all solutions with polynomial delay, even if we add lexleader constraints in LEX.

First, we introduce some more notation. Let $P$ be a class of CSPs. The class $P_{\text{LEX}}$ then consists of instances $C$ which we can decompose into two instances $C_P$ and $C_{\text{LEX}}$ such that $C_P \in P$ and $C_{\text{LEX}} \in \text{LCSP}$ with the property that the constraints in $C_{\text{LEX}}$ are constructed for a subset of the symmetries of $C_P$. We use the natural order on the domains as a value order for constructing the lexleader constraints and we adapt the variable order in the lexleader constraints, depending on $C_P$. We note that an instance $C$ has at least one solution per orbit of solutions in $C_P$ because $C_{\text{LEX}}$ only contains lexleader constraints for (some of) the symmetries of $C_P$.

We note that some tractable classes are conservative, i.e. closed under arbitrary domain reductions. So, the oracle surely runs in polynomial time. For conservative classes Theorem 7 shows that adding lexleader constraints in LEX preserves polynomial delay. This is our main contribution, which we now state explicitly.

**Corollary 8.** *Let $P$ be a class of CSPs with polynomial delay. If $P$ is conservative, then $P_{\text{LEX}}$ has polynomial delay.*

If we limit the number of lexleader constraints to be polynomial in the size of the input CSP, then we have shown the following.

**Corollary 9.** *Let $P$ be a class of CSPs with polynomial delay. If $P$ is conservative and if the size of $P_{\text{LEX}}$ is bounded by a polynomial in the size of $P$, then $P_{\text{LEX}}$ has a delay that is polynomial in the size of $P$.*

If we employ lexleader constraints for all appropriate symmetries, which may be a possibly super-exponential number, then Corollary 8 states that polynomial delay is guaranteed in the size of the CSP and the size of the (possibly super-exponentially many) lexleader constraints. Corollary 9 is geared towards practical needs. In practice, we typically limit the number of lexleader constraints to be at most polynomial by only selecting a subset of the symmetries. Corollary 9 then guarantees that polynomial delay is preserved in the size of the original instance – this is remarkable because a polynomial sized set of lexleader constraints may exclude super-exponentially many solutions.

Many well-known classes of CSPs with polynomial delay are conservative. Examples include CSPs with constraints in a tractable conservative language (Bulatov, 2003), CSPs with a perfect microstructure (Salamon and Jeavons, 2008), CSPs with the broken triangle property (Cooper et al., 2010) and CSPs with fixed bounded degree of cyclicity (Jeavons et al., 1994).

# 6 Conclusion

We have shown that for a conservative and tractable class of CSPs $P$, we can still find all solutions with polynomial delay after adding lexleader constraints in the class LEX to each instance in $P$. If we only add a polynomial number of certain lexleader constraints to the instances in $P$, we have shown that the delay is polynomial also in the size of $P$. Our results mean that adding certain lexleader constraints enables us to find all symmetrically distinct solutions in less time.

Our results highlight the necessity to use the same orders for search and for lexicographic ordering. This has been noticed empirically before and our results provide a theoretical support for this.

**Acknowledgements.** The first author thanks Dave Cohen and Andras Salamon for discussions on the topic. He is supported by the Embark initiative of the Irish Research Council for Science, Engineering and Technology.